\documentclass{article}
\usepackage{spconf,amsmath,graphicx,xcolor}
\usepackage{array, url}

\DeclareMathOperator*{\softmax}{softmax}

\newcommand\textss[1]{\textsuperscript#1}

\newcommand{\karen}[1]{\textcolor{blue}{#1 --KL}}
\newcommand{\bowen}[1]{\textcolor{magenta}{#1 --BS}}
\newcommand{\greg}[1]{\textcolor{red}{#1 --GS}}
\newcolumntype{C}[1]{>{\centering\arraybackslash}p{#1}}

 \renewcommand{\karen}[1]{}
 \renewcommand{\bowen}[1]{}
 \renewcommand{\greg}[1]{}

\title{American Sign Language fingerspelling recognition \\ in the wild}
\name{Bowen Shi$^1$, Aurora Martinez Del Rio$^2$, Jonathan Keane$^2$, Jonathan Michaux$^1$\\ \em Diane Brentari$^2$, Greg Shakhnarovich$^1$, and Karen Livescu$^1$}

\address{\textss{1}Toyota Technological Institute at Chicago, USA\hspace{0.6in} \textss{2}University of Chicago, USA\\
  {\small \tt {\{bshi, jmichaux, greg, klivescu\}@ttic.edu, \{amartinezdelrio, dbrentari,jonkeane\}@uchicago.edu}}}

\begin{document}
%
\maketitle
\begin{abstract}
We address the problem of American Sign Language fingerspelling recognition ``in the wild'', 
using videos collected from websites.
We introduce the largest data set available so far for the problem of fingerspelling recognition, and the first using naturally occurring video data.  Using this data set, we present the first attempt to recognize fingerspelling sequences in this challenging setting.
Unlike prior work, our video data is extremely challenging due to low frame rates and visual variability.  To tackle the visual challenges,
we train a special-purpose
signing hand detector using a small subset of our data.  Given the
hand detector output, a sequence model decodes the hypothesized
fingerspelled letter sequence.  For the sequence model, we explore
attention-based recurrent encoder-decoders and CTC-based approaches.
As the first attempt at fingerspelling recognition
in the wild,
this work is intended to serve as a baseline for future work on sign
language recognition in realistic conditions.  We find that, as
expected, letter error rates are much higher than in previous work on
more controlled data, and we analyze the sources of
error and effects of model variants.
\end{abstract}
\begin{keywords}
American Sign Language, fingerspelling, connectionist temporal classification, attention models
\end{keywords}

\section{Introduction}
\label{sec:intro}
\vspace{-.05in}

Sign languages, consisting of sequences of grammatically structured handshapes and gestures, is a chief means of communication among deaf people around the world.\footnote{In some settings there is a cultural distinction between the terms ``deaf'' and ``Deaf''.  In this paper, we use the term ``deaf'' to refer to both.}  \karen{added footnote}
In the US, American Sign Language (ASL) is the primary language for about 350,000 to 500,000 deaf people~\cite{remitchell} and is used by many others as a second language.
Automatic recognition of sign language would help facilitate communication between deaf and hearing individuals.  It could also enable services such as search and retrieval in deaf social and news video media, which often has little or no text associated with it.

A number of challenges are involved in sign language recognition.  Sign language employs multiple elements such as handshapes, arm movement and facial expressions.  All of these gestures are subject to coarticulation and phonological effects, so they often do not appear in their canonical forms.  In addition, there is a great deal of variability in the appearance of different signers' hands and bodies.  Finally, the linguistics of sign language is less well studied than that of spoken language, and there is much less annotated data than there is for spoken languages.

\begin{figure}
\centering
\includegraphics[width=\linewidth]{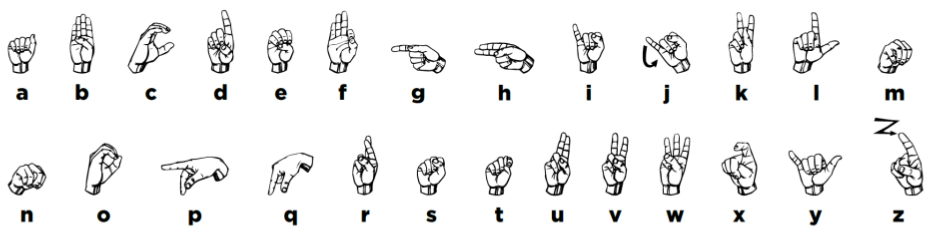}
\vspace{-0.15in}
\caption{The ASL fingerspelling alphabet, reproduced from \cite{jkean}.}
\vspace{-0.15in}
\label{fig:alphabet}
\end{figure}

In this paper we focus on the recognition of ASL fingerspelling, a component of ASL in which words are signed by a series of handshapes or trajectories corresponding to single letters (using the English alphabet). The ASL fingerspelling alphabet is shown in Figure \ref{fig:alphabet}. Fingerspelling is mainly used for lexical items that do not have their own ASL signs, such as proper nouns or technical terms, which are often important content words.  Overall, fingerspelling accounts for 12--35\% \cite{padden} of ASL, and appears frequently in technical language, colloquial conversations involving names, conversations involving current events,
emphatic forms, and the context of codeswitching between ASL and English \cite{multi_origin,padden2,code_mixing}.  Transcribing even only the fingerspelled portions of videos in online media could add a great deal of value, since these portions are often dense in content words.

Compared to sign language recognition in general, fingerspelling recognition is in some ways more constrained because it involves a limited set of handshapes, and in ASL it is produced with a single hand (unlike in some other sign languages), which makes hand occlusion less problematic \cite{pugeault}.\footnote{Two-handed fingerspelling occasionally occurs, including in our data.}  On the other hand, fingerspelling recognition presents its own challenges.  It involves very quick, small motions that can be highly coarticulated.  In lower-quality video, motion blur can be very significant during fingerspelled portions.

Most publicly available sign language data sets have been
collected in a studio or other carefully controlled environment.
Collecting such data is expensive and time-consuming, and as a result
most existing sign language data sets are fairly small.  On the other
hand, there are large amounts of fingerspelling (and, more generally,
sign language) video available online on deaf social media and news sites (e.g., \texttt{deafvideo.tv, aslized.org}).  

In this paper we focus on recognition of fingerspelling occurring in online videos, which allow us to study the recognition problem in a more natural and practical setting than in previous work.  This work is to our knowledge the first attempt at fingerspelling recognition (or any sign language recognition) ``in the wild'', that is in naturally occurring video.  For this purpose we use a newly collected data set (see Section~\ref{sec:data}), which we make available publicly.\footnote{The data set is available for download from \texttt{http://ttic.edu/livescu/chicago-fingerspelling- in-the-wild}.}
Collecting and annotating such naturally occurring video clips involves some effort, but it is much quicker to obtain a large quantity of video from a large variety of signers than in studio data collection.  

Our recognizer consists of a hand detector trained to detect the signing hand, whose output (the cropped signing hand image) is fed to an end-to-end neural sequence model.  As expected, we find that our new data set is challenging, leading to accuracies that are significantly lower than previously reported results on studio-based data sets~\cite{bshi} when using similar models.  We explore a number of neural sequence models including encoder-decoder and connectionist temporal classification (CTC)-based models. 

Our experiments show the importance of the signing hand detector for obtaining high-resolution regions of interest, and that CTC-based models outperform encoder-decoder models on our task.  We analyze the sources of error and the effect of a number of design choices.

\vspace{-.05in}
\section{Related work}
\vspace{-.05in}
There has been a significant amount of work on automatic sign language recognition.
Video corpora have been collected for a variety of sign languages~\cite{agris,athitsos,jhuang}.
These data sets are all recorded in a studio environment,
which makes the variability lower than in natural day-to-day signing.  One example of a more naturalistic data set is the RWTH-PHOENIX-Weather corpus \cite{rwth}, which contains German Sign Language in the context of daily television weather forecasts;
however, the number of signers is still limited (7 signers) and the visual variability is fairly controlled.  Fingerspelling-specific data sets are much rarer. The ChicagoFSVid data set is the largest of which we are aware; it includes 4 native ASL signers fingerspelling 600 sequences each, and has been used in recent work on lexicon-free recognition and signer adaptation~\cite{tkim3,bshi}.  The National Center for Sign Language and Gesture Resources (NCSLGR) Corpus includes about 1,500 fingerspelling sequences (as well as a variety of other ASL signs)~\cite{ncslgr,neidle-tool}.  In addition to video, many efforts have been devoted to using depth sensors instead of or in addition to video, which can be very helpful for developing new interfaces~\cite{pugeault,greek_sl}.  In this work, however, we focus on naturally occurring online data, which is typically in the form of video.

Automatic sign language recognition can be approached similarly to speech recognition, with image frames and signs being treated analogously to audio signals and words or phones respectively.
As in a number of other domains, convolutional neural networks (CNNs) have recently been replacing engineered features in sign language recognition research \cite{deephand,resign,rcn,jhuang,bshi}.  For sequence modeling, most previous work has used hidden Markov models (HMMs) \cite{marasgos:phmm,deephand,resign,tkim3}, and some has used segmental conditional random fields \cite{tkim1,tkim2,tkim3}. Much of this work relies on frame-level labels for the training data.
Due to the difficulty of obtaining frame-level annotation, recent work has increasingly focused on learning from sequence-level labels alone~\cite{deephand,jhuang,rcn}.

Specifically for fingerspelling recognition, most prior work has focused on restricted settings.
When the lexicon is restricted to a small size (20 - 100 words), letter error rates lower than 10\% have been achieved \cite{pgoh,sliwicki,sricco}.  Another important restriction is the signer identity.
In \cite{tkim3,bshi}, letter error rates of less than 10\% were achieved in a lexicon-free setting (unrestricted vocabulary) when training and testing on the same signer, but the error rate increases to above 40\% in the signer-independent setting. The large performance gap between these two settings
has also been observed in general sign language recognition \cite{rwth}.

Most fingerspelling recognition approaches begin by extracting the signing hand from the image frames~\cite{tkim1,tkim3,jhuang}.  
Due to the high quality of video used in prior work, hand detection (or segmentation) is usually treated as a pre-processing step with high accuracy, with little analysis of its impact on performance.
In our new data set, the variation in hand appearance, motion blur, and backgrounds makes the hand extraction problem much more challenging.

\vspace{-.05in}
\section{Data}
\label{sec:data}
\vspace{-.05in}
We have collected a new data set consisting
of fingerspelling clips from ASL videos on
YouTube, \texttt{aslized.org} and \texttt{deafvideo.tv}. ASLized
is an organization that creates educational videos that pertain to the use, study, and structure of ASL. DeafVideo.tv is a social media website for deaf vloggers, where users post videos on a wide range of topics.  The videos include a variety of viewpoints and styles, such as webcam videos and lectures.
214 raw ASL videos were collected, and all fingerspelling clips
within these videos were manually located and annotated.

The videos were annotated by in-house annotators at TTIC and the U.~Chicago Sign Language Linguistics lab, using the ELAN video annotation tool~\cite{elan} Annotators viewed the videos, identified instances of fingerspelling within these videos, marked the beginning and end of each fingerspelling sequence, and labeled each sequence with the letter sequence being fingerspelled.  No frame-level labeling has been done; we use only sequence-level labels.
Annotators also marked apparent misspellings 
and instances of fingerspelling articulated with two hands.
The fingerspelled segments include proper nouns, other words, and abbreviations (e.g., N-A-D for National Association of the Deaf).  
The handshape vocabulary contains the 26 English letters and the 5
special characters \{$<$sp$>$, \&, \emph{'},\emph{.},\emph{@}\}
that occur very rarely.

We estimate the inter-annotator agreement on the label sequences to be about 94\%, as measured for two annotators who both labeled a small subset of the videos; this is the letter accuracy of one annotator, viewing the other as reference.

\begin{figure}[t]
\centering
\vspace{-0.25in}
\includegraphics[width=\linewidth]{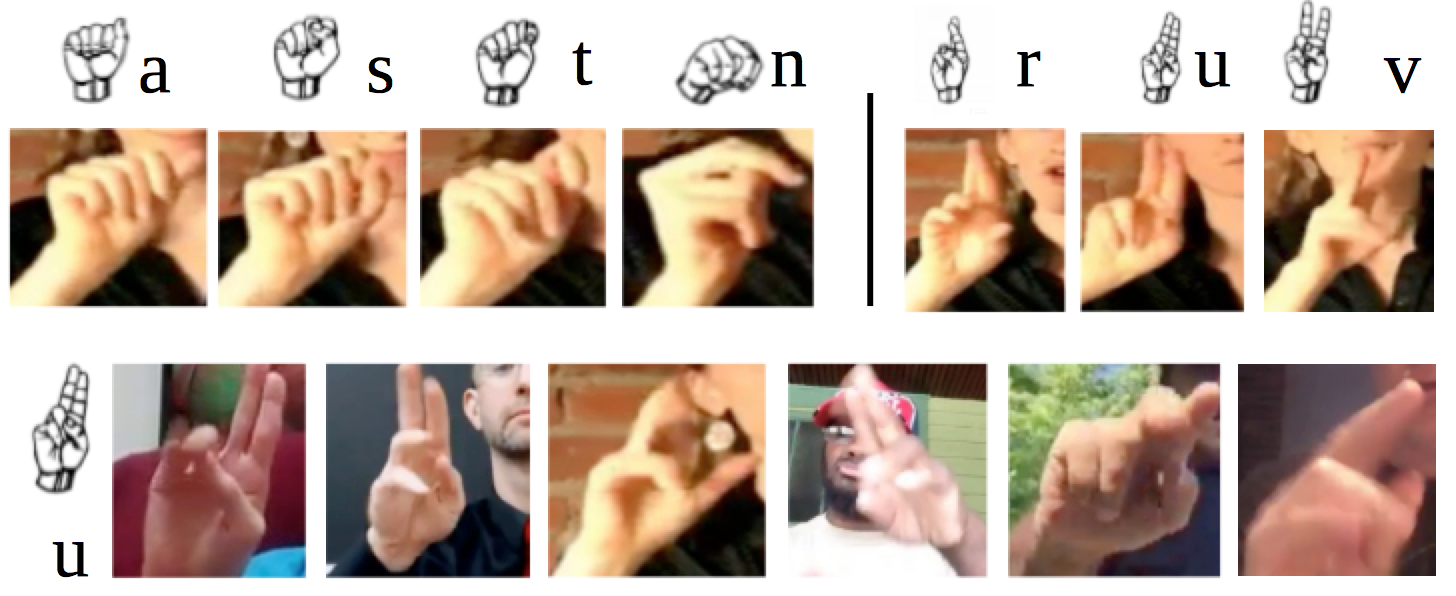}
\vspace{-0.25in}
\caption{Illustrations of ambiguity in fingerspelled handshapes.  Upper row: different letters with similar handshapes, all produced by the same signer.  Lower row: the same letter (u) signed by different signers.}
\label{fig:conf}
\vspace{-0.20in}
\end{figure}

\begin{figure}[htp]
\centering
\vspace{-0.25in}
\includegraphics[width=\linewidth]{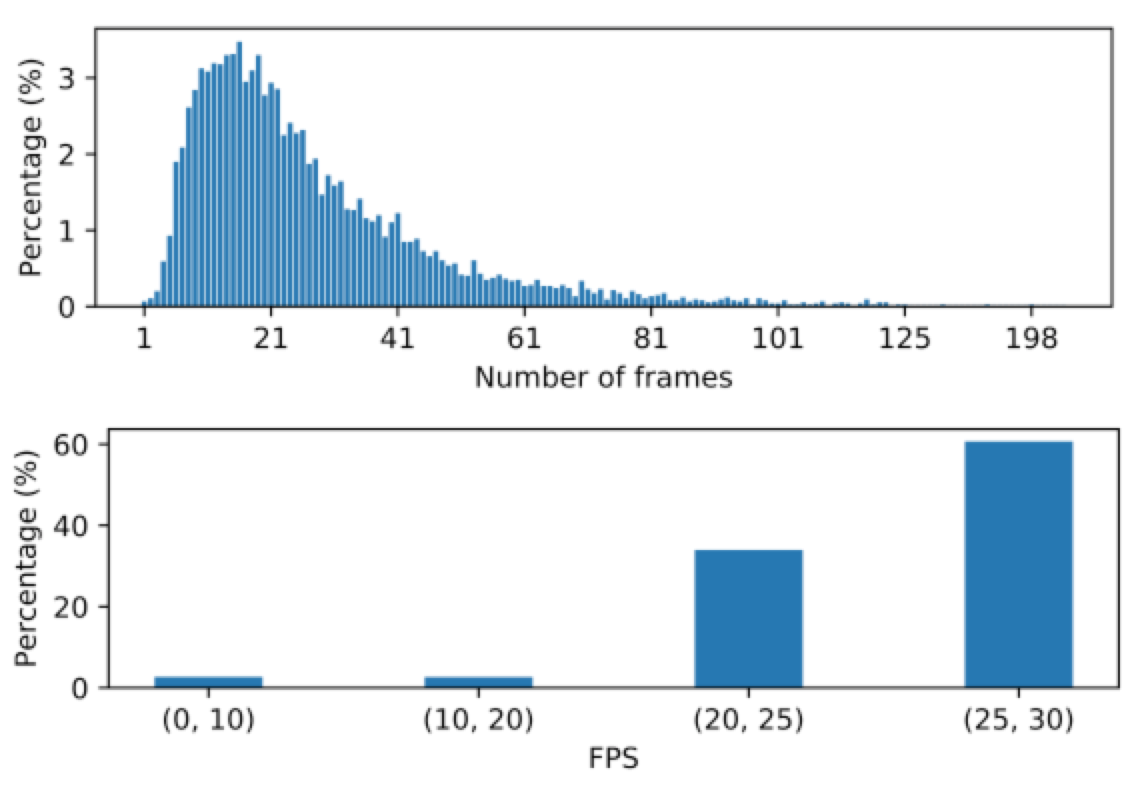}
\vspace{-0.25in}
\caption{\label{fig:statistics}Histograms of the number of frames per fingerspelled sequence and frames per second (FPS) for fingerspelled sequences in the data set.}
\vspace{-0.20in}
\end{figure}

As a pre-processing step,
we removed all fingerspelling video sequences containing fewer frames
than the number of labels.  We split the remaining data (7304
sequences) into 5455 training sequences, 981 development (dev) sequences, and 868 test sequences.  
Using frames per second (FPS) as a proxy for video quality, we ensured that the distribution of FPS was roughly the same across the three data partitions.  The dataset includes about 168 unique signers (91 male, 77 female).\footnote{These numbers are estimated by visual inspection of the videos, as most do not include meta-data about the signer.}  192 of the raw videos contain a single signer, while 22 videos contain multiple people.
Each unique signer is assigned to only one of the data
partitions.  The majority of the fingerspelling sequences are
right-handed (6782 sequences), with many fewer being left-handed (522
sequences) and even fewer two-handed (121 sequences).  Roughly half of
the sequences come from spontaneous sources such as blogs and
interviews; the remainder comes from scripted sources such as news,
commercials, and academic presentations. The frame resolution has a
mean and standard deviation of $640\times 360\,\pm\,290\times 156$. Additional statistics are
given in Figure \ref{fig:statistics}.

This data set collected ``in the wild'' poses serious challenges, such as great visual variability (due to lighting, background, camera angle, recording quality) and signing variability
(due to speed and hand appearance).
To illustrate some of these challenges, Figure~\ref{fig:conf} shows a number of representative frames from our data set.
There can be a great deal of variability in fingerspelling the same letter, as illustrated in the bottom row of Figure~\ref{fig:conf}.  In addition, many fingerpselled letters have similar handshapes.  For example, the letters \emph{a}, \emph{s}, \emph{t} and \emph{n} are only distinguished by the position of the thumb, and the letters \emph{r}, \emph{u} and \emph{v} are all signed with the index and middle fingers extended upward.  The small differences among these letters can be even harder to detect in typical lower-quality online video with highly coarticulated fingerspelling, as seen in the top row of Figure~\ref{fig:conf}.

\vspace{-.05in}
\section{Model}
\vspace{-.05in}

\begin{figure}[htp]
\centering
\vspace{-0.15in}
\includegraphics[width=\linewidth]{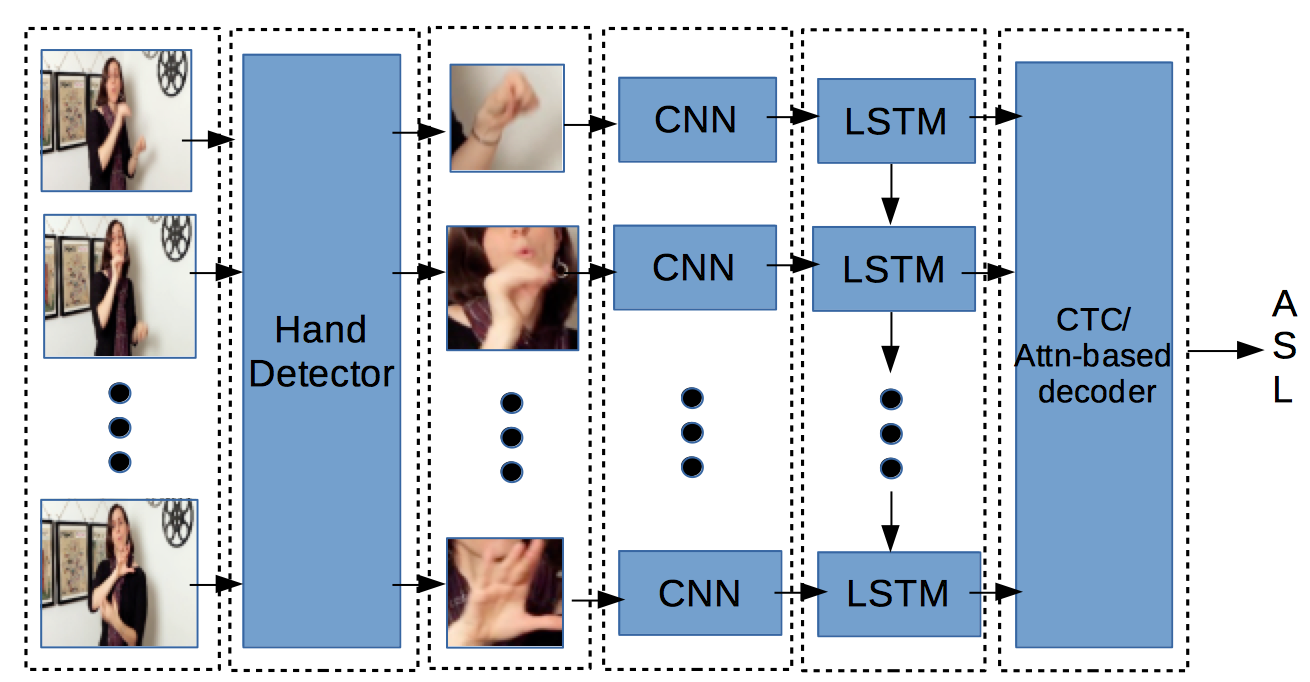}
\vspace{-0.15in}
\caption{\label{fig:model}Sketch of our approach.  After the hand detector component, the rest of the model is trained end-to-end.}
\vspace{-0.25in}
\end{figure}

Our approach to fingerspelling recognition consists of a signing hand detector followed by a sequence recognizer, illustrated in Figure~\ref{fig:model}.

\vspace{-.05in}
\subsection{Signing hand detection}
\label{sec:tube_gen}
The hand detection problem here is somewhat different from typical
hand detection.
\karen{changed ``Most'' to ``A large proportion''.  But if you've measured it and ``Most'' is correct, we can change it back}
A large proportion of the video frames contain more than one hand,
but since ASL fingerspelling generally involves a single hand, the
objective here is to detect the {\it signing hand}.  This can be
viewed as a problem of action localization~\cite{action_localization}.
As in prior work on action localization \cite{action_localization}, we
train a detection network that takes as input both the image
appearance and optical flow, represented as a motion vector for every
pixel computed from two neighboring frames
\cite{opt_flow}.\footnote{For optical flow we use the OpenCV
  implementation of~\cite{Farneback}.}  For the detection
network, we adapt the design of the Faster R-CNN object detector
\cite{faster_rcnn}. As in \cite{faster_rcnn}, the detector is based on
an ImageNet-pretrained VGG-16 network \cite{imagenet,vgg}.
Unlike the general object detector, we only preserve the first 9
layers of VGG-16 and the stride
of the network is reduced to 4. Lower layers able to capture more fine
details~\cite{visualization} combined with finer stride/localization
are beneficial for detecting signing hands, which tend to be small
relative to the frame size.

Unlike much work in action localization \cite{action_localization},
which processes optical flow and appearance images in two distinct
streams, we concatenate the optical flow and RGB image as the input to
a single CNN. In our video data, motion involves many
background objects like faces and non-signing hands, so a separate optical flow stream may be misleading.

Given bounding boxes predicted framewise by the Faster R-CNN, we first
filter them by spatial non-maxima
suppresion (NMS)~\cite{girshick:fast_rcnn}, greedily removing any box
with high overlap with a higher-scoring box in the same frame. Next, we
link the surviving boxes across time to form a video region likely to be associated with a fingerspelling sequence,
which we call a ``signing tube'' (analogously to action tubes in
action recognition~\cite{malik:action_tube}). Even after NMS, there
may be multiple boxes in a single frame (e.g., when the signer is
signing with both hands). Our temporal linking process helps prevent
switching between hands in such cases. It also has a smoothing effect, which can
reduce errors in prediction compared to that based on a single frame.

More formally, the input to the signing tube prediction is a sequence of sets of bounding box coordinate and score pairs: $\{(b_t^1, s_t^1), (b_t^2, s_t^2), ..., (b_t^n, s_t^n)\}$, $1\leq t\leq T$, produced by the frame-level signing hand detector. The score $s_t^i$ is the probability of a signing hand output by the Faster R-CNN. We define the {\it linking score} of two boxes $b_t^i$ and $b_{t+1}^j$ in two consecutive frames as:
\begin{equation}
e(b_t^i, b_{t+1}^j)=s_t^i+s_{t+1}^j+\lambda * IoU(b_t^i, b_{t+1}^j)
\end{equation}
\noindent where $IoU(b_t^i, b_{t+1}^j)$ is the intersection over union of $b_t^i$ and $b_{t+1}^j$ and $\lambda$ is a hyperparameter that is tuned on held-out data.
Generation of the optimal signing tube is the problem of finding a sequence of boxes $\{b_1^{l_1}, ..., b_T^{l_T}\}$ that maximizes the sequence score, defined as
\begin{equation}
E(l)=\frac{1}{T}\sum_{t=1}^{T-1}e(b_t^{l_t}, b_{t+1}^{l_{t+1}})
\end{equation}
\noindent This optimization problem is solved via a Viterbi-like dynamic programming algorithm \cite{vtb}. 

\vspace{-.05in}
\subsection{Fingerspelling sequence model}
We next take the signing tube, represented as a sequence of image patches $\{\mathbf{I}_1, \mathbf{I}_2,..., \mathbf{I}_T\}$, as input to a sequence model that outputs the fingerspelled word(s) $w$.  We work in a lexicon-free setting, in which the word vocabulary size is unlimited, and represent the output $w$ as a sequence of letters $w_1, w_2, ..., w_s$.  The model begins by applying several convolutional layers to individual image frames to extract feature maps.  
The convolutional layers transform the frame sequence $\{\mathbf{I}_1, \mathbf{I}_2,..., \mathbf{I}_T\}$ into a sequence of features $\{\mathbf{f}_1, \mathbf{f}_2,..., \mathbf{f}_T\}$.

The sequence of image features $\{\mathbf{f}_1, \mathbf{f}_2,..., \mathbf{f}_T\}$ is then fed as input to a long short-term memory recurrent neural network (LSTM)~\cite{lstm} that models the temporal structure, producing a sequence of hidden state vectors (higher-level features) $\{\mathbf{e}_1, \mathbf{e}_2,..., \mathbf{e}_T\}$.\karen{rephrased}  Given the
features produced by the LSTM, the next step is to compute the probabilities of the letter sequences $w_1, w_2, ..., w_s$. \karen{made notation consistent with paragraph above}
We consider two approaches for decoding, neither of which requires
frame-level labels at training time: an attention-based LSTM decoder,
and connectionist temporal classification (CTC)~\cite{ctc}.  In the former case, the whole sequence model becomes a recurrent encoder-decoder with attention~\cite{attn}.

In the \textbf{attention-based model}, temporal attention weights are applied to $(\mathbf{e}_1, \mathbf{e}_2,..., \mathbf{e}_T)$ during decoding, which allows the decoder to focus on certain chunks of visual features when producing each output letter.  If the hidden state of the decoder LSTM at timestep $t$ is $\mathbf{d}_t$, the probability of the output letter sequence is given by \karen{edited for consistency, and also edited the 3rd and 4th equations below}
\begin{equation}
\begin{split}
& \alpha_{it} = \softmax(\mathbf{v}_d^T \tanh(\mathbf{W}_e\mathbf{e}_i + \mathbf{W}_d \mathbf{d}_t)) \\
& \mathbf{d}_t^\prime = \displaystyle\sum_{i=1}^T \alpha_{it}\mathbf{e}_i \\
& p(w_t|w_{1:t-1}, \mathbf{e}_{1:T})=\softmax (\mathbf{W}_o[\mathbf{d}_t;\mathbf{d}_t^\prime]+\mathbf{b}_o) \\
& p(w_1, w_2, \ldots, w_s|\mathbf{e}_{1:T})=\displaystyle\prod_{t=1}^s{p(w_t|w_{1:t-1}, \mathbf{e}_{1:T})} \\
\end{split}
\end{equation}
\noindent where $\mathbf{d}_t$ is given by the standard LSTM update equations \cite{lstm}.  The model is trained to minimize log loss.

In the \textbf{CTC-based model}, \karen{edited for consistency} for an input sequence of $m$-dimensional visual feature vectors $\mathbf{e}_{1:T}$, we define a continuous map $\mathcal{N}_w: (\mathcal{R}^m)^T \mapsto (L^\prime)^T$ representing the transformation from $m$-dimensional
features $\mathbf{e}_{1:T}$ to frame-level label probabilities and a many-to-one map $\mathcal{B}: {{L^\prime}^T} \mapsto L^{\leq T}$ where $L^{\leq T}$ is the set of all possible labelings. Letting $L$ be the output label vocabulary,
$L^\prime = L\cup \{blank\}$, and $y_k^t$ the probability of
label $k$ at time $t$, the posterior probability of any labeling $\pi\in {L^\prime}^T$ is
\begin{equation}
p(\pi|\mathbf{e}_{1:T}) = \displaystyle\prod_{t=1}^T{y_{\pi_t}^t}=\displaystyle\prod_{t=1}^T {\textrm{softmax}_{\pi_t}(\mathbf{A}^e\mathbf{e}^t
+\mathbf{b}^e)}
\end{equation}
At training time, the probability of a given labeling $w = w_1, w_2, ..., w_s$
is obtained by summing over all the possible frame-level labelings $\pi$,
which can be computed by a forward/backward algorithm:
\begin{equation}\label{eq:ctc_s2}
p(w|\mathbf{e}_{1:T})=\displaystyle\sum_{\pi\in\mathcal{B}^{-1}(w)}p(\pi|\mathbf{e}_{1:T})
\end{equation}
\noindent The CTC model is trained to optimize this probability for the ground-truth label sequences.

Finally, in decoding we combine these basic sequence models with an RNN language model.  To decode with a language model, we use beam search to produce several candidate words at each time step and then rescore the hypotheses in the beam using the summed score of the sequence model, weighted language model, and an insertion penalty to balance the insertion and deletion errors. The language model weight and insertion penalty are tuned.

\vspace{-.05in}
\section{Experiments}
\vspace{-.05in}

All of the experiments are done in a signer-independent, lexicon-free (open-vocabulary) setting using the data set and partitions described in Section~\ref{sec:data}.

\textbf{Evaluation} We measure the letter accuracy of predicted sequences, as is commonly used in sign language recognition and speech recognition:
$Acc = 1 - \frac{S+I+D}{N}$, 
where S, I and D are the numbers of substitutions, insertions, and deletions (with respect to the ground truth) respectively, and N is the number of letters in the ground-truth transcription. 

\textbf{Hand detection details}
We manually annotated every frame in 180 video clips from our training set \karen{added} with signing and non-signing hand bounding boxes.\footnote{The non-signing hand category is annotated in
  training to help the detector learn the distinction between signing
  hands, other hands, and background; once the detector is trained we
  ignore the non-signing hand category, and only use the signing hand detections.}
Of these, 123 clips (1667 frames) are used for training and 19 clips (233 frames) for validation.  All images are resized to $640\times 368\times 3$.  We use stochastic gradient descent (SGD) for optimization, with initial learning rate 0.001
and decreased by a factor of $2$ every 5 epochs.
We apply greedy per-frame NMS with intersection-over-union (IoU)\footnote{IoU is
the ratio of the area of overlap over area of union of two regions.}
threshold of 0.9, until 50 boxes/frame remain. The bounding boxes are then smoothed as described in Section~\ref{sec:tube_gen}. $\lambda$ is tuned to 0.3, which maximizes the proportion of validation set bounding boxes with IoU $> 0.5$.  
Using our bounding box smoothing approach, the proportion of bounding
boxes with IoU $> 0.5$ is increased from 70.0\% to 77.5\%.\greg{what set?}

\textbf{Letter sequence recognition details} The input to the recognizer is a bounding box of the predicted signing hand region. All bounding boxes are resized to $224\times 224$ before being fed to the sequence model.
For the convolutional layers of the sequence model, we use AlexNet \cite{alexnet} pre-trained on ImageNet as the base architecture.\footnote{A deeper network like VGG cannot be used due to the memory requirements introduced by its small stride.}   For the recurrent network, we use a single-layer long short-term memory (LSTM) network with 600 hidden units. (A model with more recurrent layers does not consistently improve performance.)  The network weights are learned using mini-batch stochastic gradient descent (SGD) with weight decay. The initial learning rate is 0.01 and is decayed by a factor of 10 every 15 epochs. Dropout with a rate of 0.5 is used between fully connected layers of AlexNet.  The batch size is 1 sequence in all experiments. The hyperparameters were tuned to maximize the dev set letter accuracy.
The language model is a single-layer LSTM with 600 hidden units, trained on the letter sequences in our training set.

\vspace{-.05in}
\subsection{Main results}
\label{sec:main_baseline}
\vspace{-.05in}
Table \ref{tab:res} shows the performance of our models (``Hand'') using the cropped hand region as input to the sequence model, as well as of a baseline model (``Global'') with the same sequence model architecture but without hand detection (i.e., taking the whole image as input).  This baseline model is based on commonly used approaches for video description~\cite{conv_lstm}.
For the Global baseline, image frames are resized to $224\times 224$ due to memory constraints.  We also report the result of a ``guessing'' baseline (``LM'') that predicts words directly from our language model.
This baseline
only uses statistics of fingerspelled letter sequences, uses no visual input for prediction, and always predicts the
output of a greedy decoding of the language model.

The Global baseline outperforms the language model baseline by a small margin, suggesting that the full-image model is unable to use much of the visual information.
Compared to the baseline, our approach with hand detection performs much better.  The hand detection step both filters out irrelevant information (e.g. background, non-signing hand) and allows us to use higher resolution image regions.
CTC-based models consistently outperform the encoder-decoder models on this task.
Since fingerspelling sequences are expected to have largely monotonic alignment with the video, this may benefit the simpler CTC model.

\textbf{Human performance}
Although we do not have a precise measure of human performance,\footnote{Inter-annotator agreement is not a good measure, since the annotators see the entire video surrounding each fingerspelled sequence, while the automatic recognizers see only the fingerspelled sequence.} we estimated it informally in the following way.  We took a small set of (153) fingerspelling sequences that were located and labeled by one annotator who had access to the full videos as usual.  Another annotator then labeled the fingerspelling sequences only, without access to the surrounding video.  Relative to the first annotator, the second annotator had a letter accuracy of 82.7\%.  We do not have this measure on our full test set, and did not carefully control for the order of presentation of the data to the annotator; but this provides a rough idea of the difficulty of the task for humans.  

\begin{table}[t]
\centering
\vspace{-0.15in}
\caption{Letter accuracies (\%) for the language model and global baseline models and our hand detector-based models.}
\label{tab:res}
\begin{tabular}{l||C{0.05\linewidth}|C{0.12\linewidth}|C{0.1\linewidth}|C{0.1\linewidth}|C{0.1\linewidth}}\hline
 & \small{LM} & \begin{tabular}{@{}c@{}}Global \\ enc-dec\end{tabular} & \begin{tabular}{@{}c@{}}Global \\ CTC\end{tabular} & \begin{tabular}{@{}c@{}}Hand\\ enc-dec\end{tabular} & \begin{tabular}{@{}c@{}}Hand\\ CTC\end{tabular}\\ \hline
Test Acc (\%)& 9.4 & 12.7 & 10.0 & 35.0 & 41.9 \\ \hline
\end{tabular}
\vspace{-0.15in}
\end{table}

\textbf{Additional model variants}
Besides the proposed model, we also considered a number of variants that we ultimately rejected.  The bounding boxes output by the hand detector fail to contain the whole hands in many cases.  We considered enlarging the predicted bounding box by a factor of $s$ in width and height before feeding it to the recognizer. 
In addition, we also considered using optical flow as an additional input channel to the sequence model (in addition to the hand detector), since motion information is important in our task.

We find that neither of these variants consistently and/or significantly improves
performance (on the dev set) compared to the baseline with $s=1$
and no optical flow input. 
Thus we do not pursue these model variants further.

\textbf{Error analysis}
The most common types of errors are deletions, followed by insertions.  The encoder-decoder model makes more insertion errors and fewer deletion errors than the CTC model, that is its error types are more balanced, but its overall performance is worse.  The most common substitution pairs for the CTC model are (u $\rightarrow$ r), (o $\rightarrow$ e), (y $\rightarrow$ i), (w $\rightarrow$ u), (y $\rightarrow$ i) and (j $\rightarrow$ i) (see Table~\ref{tab:sub_error}).\karen{added table ref}
 (u $\rightarrow$ r), (y $\rightarrow$ i) and (w $\rightarrow$ u) involve errors with infrequent letters, which may be due to the relative dearth of training data for these letters.  The pair (j $\rightarrow$ i) is interesting in that the most important difference between them is whether the gesture is dynamic or static.  Compared to studio data, the frame rates in our data are much lower, which may make it more difficult to distinguish between static and dynamic letters with otherwise similar handshapes.

Since deletions are the most frequent error type, applying an insertion/deletion penalty is one possible way to improve performance.  Using such a penalty produces a negligibly small improvement, as seen in Table~\ref{tab:lm}.

\begin{table}
\centering
\vspace{-0.10in}
\caption{\label{tab:sub_error}
  Percentage of several important substitution error pairs on the development set.  For a given label pair ($x_1 \rightarrow x_2$), this is the percentage of occurrences of the ground-truth label $x_1$ that are recognized as $x_2$. \karen{do these numbers reflect the new data division?}\bowen{Yes}}
\begin{tabular}{c||c|c|c|c|c}\hline
 & (u $\rightarrow$ r) & (o $\rightarrow$ e) & (y $\rightarrow$ i) & (w $\rightarrow$ u) & (j $\rightarrow$ i) \\ \hline
 \%& 17.0 & 11.7 & 7.9 & 7.6 & 6.7 \\ \hline
\end{tabular}
\vspace{-0.15in}
\end{table}

\begin{figure}[htp]
\centering
\vspace{-0.15in}
\includegraphics[width=\linewidth]{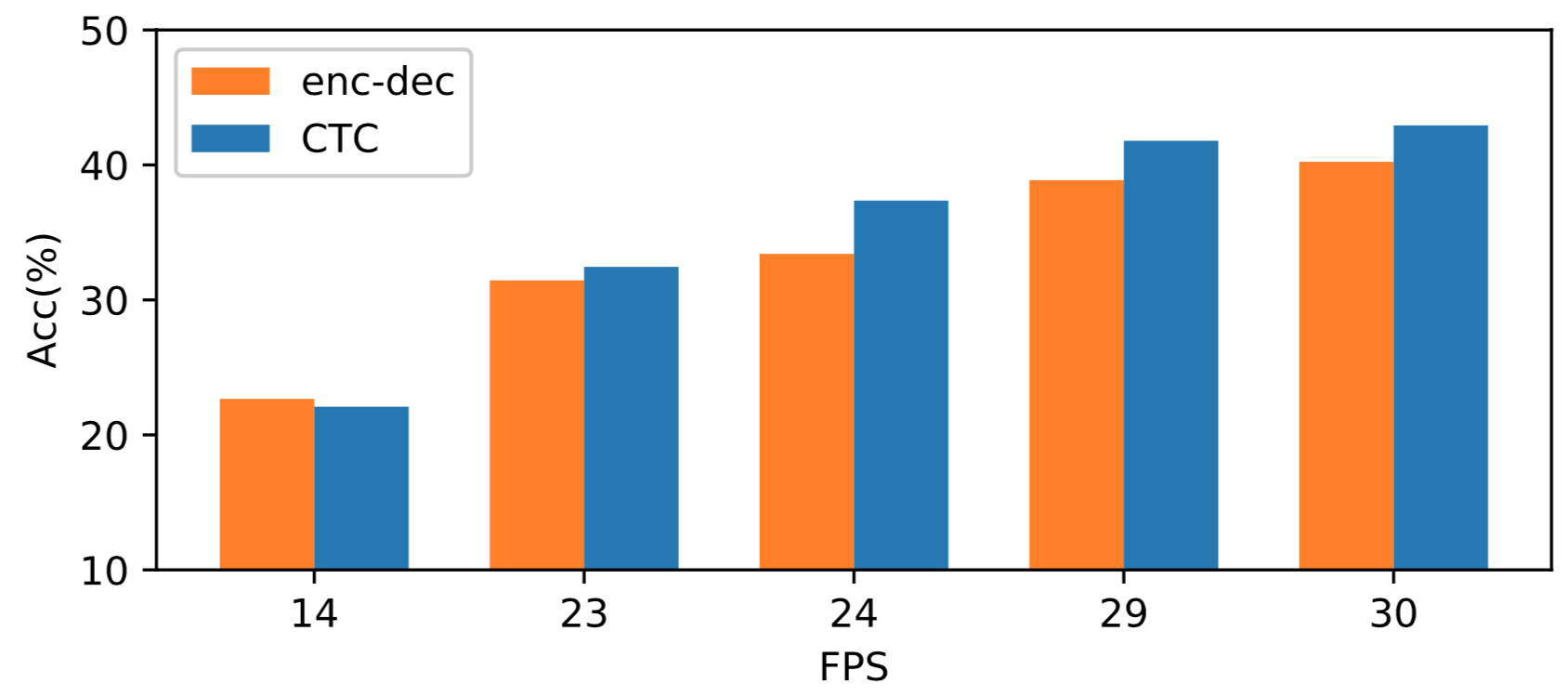}
\vspace{-0.35in}
\caption{\label{fig:fps}Development set accuracy for sequences with different frame rates (FPS) for our CTC and encoder-decoder models.}
\vspace{-0.15in}
\end{figure}

To measure the impact of video quality on performance, we divided the dev set into subsets according to the frame rate (FPS) and report the average error in each subset (see Figure~\ref{fig:fps}). In general, higher frame rate corresponds to higher
accuracy.

\textbf{Effect of the language model}
\label{sec:lm}
Next we consider to what extent the language model 
improves performance.
It is not clear how much the language model can help, or what training material is best, 
since fingerspelling does not follow the same distribution as English words and there is not a great deal of transcribed fingerspelling data available.  
In addition to training on the letter sequences in our own training set, we also consider training on all words in the CMUdict (version 0.7a) dictionary~\cite{cmu_dict}, which contains English words and common names, and no improvement was found.  The development set perplexity of our LM trained 
with in-house data is 17.3.  Since the maximum perplexity is 32 (31 characters plus end-of-sequence), this perplexity reflects the difficulty of learning the statistics of fingerpselled letter sequences.
We also experimentally check the effect of the insertion penalty and beam search.
The beam size, language model weight, and insertion penalty are tuned and the best development set results are given in Table \ref{tab:lm}.    Using a language model, the accuracy is improved by a small margin ($\sim$1\%).\karen{edited margin}

\begin{table}
\centering
\caption{\label{tab:lm} Development set letter accuracies (\%) when decoding with a language model (lm: LM trained with words from our training set, beam: beam search, ins: insertion penalty, no LM: greedy decoding).}
\begin{tabular}{c||C{0.115\linewidth}|C{0.115\linewidth}|C{0.115\linewidth}|C{0.115\linewidth}}\hline
 & \small{no LM}  &  \small{+ beam} &\begin{tabular}{@{}c@{}}\small{+ beam} \\ \small{+ ins} \end{tabular}& \begin{tabular}{@{}c@{}}\small{+ beam} \\ \small{+ ins +lm} \end{tabular} \\ \hline
 \small{CTC} & 41.1 & 41.1 & 41.4 & 42.8 \\ \hline
 \small{Enc-dec} & 35.7 & 35.8 & 35.9 & 36.7 \\ \hline
\end{tabular}
\vspace{-0.20in}
\end{table}

\vspace{-.1in}
\section{Conclusion}
\vspace{-.05in}
This work has studied for the first time the recognition of ASL fingerspelling in naturally occurring online videos.  Our newly collected data set includes a variety of challenging visual conditions.  We have seen that a purpose-built hand detector, with smoothing over time, is very helpful.
The best test set letter accuracies we obtain, using a CTC-based recognizer, are around 42\%,\karen{edited number} indicating that there is room for much future work.
Although our data set is the largest fingerspelling data set to our knowledge, it is still much smaller than typical speech recognition corpora, and we are continuing to collect additional online video data.

\vspace{-.2in}
\section*{Acknowledgements}
\vspace{-.15in}
We are grateful to our data annotators Raci Lynch and Amy Huang, and to Raci Lynch also for data collection and help with the annotation setup.  This research was funded by NSF grants 1433485 and 1409886.

\bibliographystyle{IEEEbib}
\bibliography{refs}

\end{document}